\documentclass[lettersize,journal]{IEEEtran}
\usepackage[dvipsnames]{xcolor}
\usepackage{amsmath,amsfonts,amssymb,mathtools,amsthm}
\usepackage{algorithmic}
\usepackage{algorithm}
\usepackage{array}
\usepackage[caption=false,font=normalsize,labelfont=sf,textfont=sf]{subfig}
\usepackage{textcomp}
\usepackage{stfloats}
\usepackage{url}
\usepackage{verbatim}
\usepackage{graphicx}
\usepackage{cite}
\usepackage{glossaries}
\def\BibTeX{{\rm B\kern-.05em{\sc i\kern-.025em b}\kern-.08em
		T\kern-.1667em\lower.7ex\hbox{E}\kern-.125emX}}
\usepackage{balance}
\usepackage{pgfplots}
\pgfplotsset{compat=1.18}
\newlength\axwidth
\newlength\axheight
\newlength\myfontsize
\usetikzlibrary{angles,quotes,arrows.meta,calc,positioning}
\usepackage{graphicx}
\usepackage{hyperref}
\hypersetup{colorlinks=true, linkcolor=blue, citecolor=blue, urlcolor=blue, bookmarksdepth=3}

\newacronym{eeg}{EEG}{electroencephalography}
\newacronym{egda}{EGDA}{EEG-based graph-guided Domain adaptation}
\newacronym{abci}{aBCI}{EEG-based affective brain–computer interface}
\newacronym{erp}{ERP}{event-related potential}
\newacronym{de}{DE}{differential entropy}
\newacronym{dwt}{DWT}{discrete wavelet transform}
\newacronym{plv}{PLV}{phase-locking value}
\newacronym{emd}{EMD}{empirical mode decomposition}
\newacronym{gnn}{GNN}{graph neural network}
\newacronym{rnn}{RNN}{recurrent neural network}
\newacronym{cnn}{CNN}{convolutional neural network}
\newacronym{lstm}{LSTM}{long short-term memory}
\newacronym{da}{DA}{domain adaptation}
\newacronym{mmd}{MMD}{maximum mean discrepancy}
\newacronym{gfil}{GFIL}{graph-regularized feature importance learning}
\newacronym{ogssl}{OGSSL}{optimal graph-coupled semi-supervised learning}
\newacronym{gasdsl}{GASDSL}{graph adaptive semi-supervised discriminative subspace learning}
\newacronym{srag}{SRAG}{semi-supervised regression with adaptive graph learning}
\newacronym{em}{EM}{expectation–maximization}
\newacronym{cptml}{CPTML}{coupled projection transfer metric learning}
\newacronym{jagp}{JAGP}{joint adaptation with graph propagation}
\newacronym{blc}{BLC}{brain-like computing}
\newacronym{vda}{VDA}{visual domain adaptation}
\newacronym{jsfa}{JSFA}{joint sample and feature importance assessment}
\newacronym{sragl}{SRAGL}{semi-supervised regression with adaptive graph learning}
\newacronym{sda}{SDA}{aubspace distribution alignment}
\newacronym{tca}{TCA}{transfer component analysis}
\newacronym{jda}{JDA}{joint distribution adaptation}

\begin{document}
\title{EEG-based Graph-guided Domain Adaptation for Robust Cross-Session Emotion Recognition}
\author{
	\IEEEauthorblockN{Maryam~Mirzaei, Farzaneh~Shayegh, and Hamed~Narimani}\\
	\IEEEauthorblockA{\small Department of Electrical and Computer Engineering
		Isfahan University of Technology, Isfahan, Iran}

}

\maketitle

\begin{abstract}
	
Accurate recognition of human emotional states is critical for effective human–machine interaction. Electroencephalography (EEG) offers a reliable source for emotion recognition due to its high temporal resolution and direct reflection of neural activity. Nevertheless, variations across recording sessions present a major challenge for model generalization. To address this issue, we propose \Gls{egda}, a framework that reduces cross-session discrepancies by jointly aligning marginal and class-conditional distributions, while preserving the intrinsic structure of EEG data through graph regularization. Experimental results on the SEED-IV dataset demonstrate robust cross-session
performance, achieving accuracies of 81.22\%, 80.15\%, and 83.27\% across three
transfer tasks and outperforming several baseline methods.
 Furthermore, the analysis highlights the Gamma frequency band as the most discriminative and identifies central-parietal and prefrontal brain regions as critical for reliable emotion recognition.
\end{abstract}


 \section{Introduction}
 \IEEEPARstart{A}{ccurate} recognition of human emotions is essential for developing effective human–machine interaction systems \cite{picard2000affective}. Over the past decade, numerous studies have explored emotion recognition using various sources, such as vision-based, text-based, and speech-based approaches \cite{hamzah2024eeg}. While these non-physiological methods are relatively easy to implement, they suffer from limited reliability, as individuals can deliberately disguise their facial expressions, vocal tones, or textual sentiment \cite{he2020advances}.
 	
 Among physiological signals, \gls{eeg} has emerged as one of the most informative sources for emotion recognition. \gls{eeg} directly reflects neural activity with high temporal resolution and captures the dynamics of cognitive and affective processes in the brain \cite{ding2022tsception}.
 	Despite promising progress, \gls{eeg}-based emotion recognition faces critical challenges. A major difficulty arises from the non-stationarity of \gls{eeg} signals, as their distributions vary across recording sessions, subjects, and environmental conditions \cite{jayaram2016transfer}.
 	 In particular, even for the same subject, data collected on different sessions may exhibit significant distributional shifts due to changes in mental state, electrode placement, or environmental noise \cite{jayaram2016transfer,zheng2018emotionmeter}. Consequently, models trained on one session often perform poorly when applied to another, limiting generalization. This challenge motivates the development of cross-session domain adaptation techniques that can transfer knowledge from a labeled source session to an unlabeled target session \cite{zheng2016personalizing}.
 	
 Several approaches focus on EEG-based emotion recognition using graph-based models to capture the structural relationships among samples \cite{liu2024comprehensive}. However, these models typically do not employ transfer learning to address distribution mismatches across sessions or subjects. As a result, while they effectively model inter-sample similarity, they overlook aligning feature distributions between different domains.

To address these challenges, we propose a unified framework termed \gls{egda},
which integrates domain adaptation and graph-based regularization for
cross-session EEG-based emotion recognition.
This proposed framework simultaneously mitigates discrepancies in the marginal distributions and class-conditional distributions while preserving
the intrinsic geometric and semantic structures of \gls{eeg} data.
Specifically, \gls{egda} employs an iterative pseudo-labeling strategy to
progressively refine class-level alignment and incorporates a Laplacian graph
regularizer to maintain neighborhood consistency throughout the adaptation
process.
As a result, the proposed framework enables robust and effective emotion
recognition performance across different recording sessions.

The contributions of this work are summarized as follows:
\begin{itemize}
	\item A graph-guided domain adaptation framework is proposed for cross-session EEG
	emotion recognition, in which graph regularization is explicitly integrated
	into the distribution alignment process to preserve local neighborhood
	consistency and to mitigate pseudo-label instability across sessions.
	
	\item A source-domain within-class scatter minimization constraint is introduced
	to stabilize conditional distribution alignment by enforcing compact and
	discriminative class structures, thereby reducing error propagation during
	iterative pseudo-label refinement.
	
	\item A unified subspace learning model is developed to jointly align marginal
	and class-conditional distributions, where conditional alignment is iteratively
	refined through pseudo-label updates under the joint regularization of
	neighborhood structure and within-class scatter, without requiring labeled
	target data.
\end{itemize}

 	The remainder of this paper is organized as follows. Section II reviews related works in \gls{eeg}-based emotion recognition and transfer learning. Section III presents the formulation and optimization of the proposed model. Section IV describes the experimental setup, results, and performance analysis. Finally, Section V concludes the paper.
 	
\textit{Notations}: Uppercase bold letters denote matrices, while lowercase bold
letters denote vectors. For a matrix $\boldsymbol{A}$, $\boldsymbol{a}^{i}$ and
$\boldsymbol{a}_{j}$ denote its $i$-th row and $j$-th column, respectively, and
$a_{ij}$ denotes the element at the $i$-th row and $j$-th column. The superscript
$(\cdot)^{c}$ indicates quantities associated with class $c$.


\section{Related Work}
This section reviews two categories of closely related studies: (1) EEG-based emotion recognition models evaluated on the SEED-IV dataset, and (2) transfer learning and domain adaptation approaches relevant to the proposed \gls{egda} framework. Since \gls{egda} is a domain adaptation model, both categories are essential for
positioning our contributions.

\subsection{SEED-IV--Based Emotion Recognition Methods}

\Gls{gfil} introduces a unified importance learning framework that jointly performs emotion classification and EEG feature importance estimation within a graph-regularized least-squares regression model \cite{peng2021gfil}. By incorporating an auto-weighting variable, GFIL adaptively assigns different contributions to EEG feature dimensions, enabling simultaneous feature ranking, frequency-band importance analysis, and channel selection. The graph regularization term further enforces label consistency among samples sharing the same emotional state, resulting in improved recognition accuracy and enhanced interpretability in cross-session settings.

Building upon the idea of adaptive weighting, \Gls{jsfa} proposes a joint sample and feature importance assessment framework that integrates self-paced learning with feature self-weighting into a least-squares regression formulation \cite{li2024enhanced}. By progressively emphasizing high-quality EEG samples and down-weighting noisy or ambiguous ones, JSFA improves robustness against data variability and enhances emotion recognition performance.

From a semi-supervised perspective, \Gls{ogssl} presents an optimal graph-coupled semi-supervised learning framework that jointly learns an adaptive similarity graph and infers labels for unlabeled EEG samples within a unified optimization objective \cite{peng2022ogssl}. The mutual reinforcement between graph construction and label propagation allows OGSSL to effectively exploit both labeled and unlabeled data for emotion recognition.

Similarly, \Gls{sragl} proposes a semi-supervised regression framework with adaptive graph learning for cross-session EEG-based emotion recognition \cite{sha2023semi}. By formulating emotion recognition as a semi-supervised regression problem and iteratively updating the sample similarity graph, SRAGL allows the regression model and graph structure to mutually reinforce each other, leading to improved generalization across sessions.

\Gls{gasdsl} introduces a graph-adaptive semi-supervised discriminative subspace learning framework that learns a low-dimensional discriminative representation while preserving local sample relationships through an adaptively constructed maximum-entropy graph \cite{jin2023graph}. Although GASDSL effectively integrates graph learning, subspace projection, and pseudo-label estimation, the tight coupling among these components may lead to instability, particularly in noisy or cross-session EEG scenarios.

To explicitly address cross-session distribution discrepancies, \Gls{cptml} proposes a coupled projection transfer metric learning framework that jointly performs domain alignment and graph-based metric learning \cite{shen2022coupled}. By enforcing discriminative relationships among EEG samples through intrinsic and penalty graphs while simultaneously aligning feature distributions, CPTML achieves improved cross-session emotion recognition performance. However, the coupled optimization strategy and metric learning component introduce considerable computational complexity, which may limit scalability in large-scale EEG applications.

\Gls{jagp} focuses on cross-subject EEG emotion recognition by unifying domain-invariant feature adaptation, graph-adaptive label propagation, and emotion estimation within a single objective function \cite{peng2022joint}. The graph structure is iteratively updated based on shared subspace representations, enabling label propagation and feature adaptation to mutually reinforce each other. Nevertheless, since both graph construction and feature adaptation rely on intermediate pseudo-labels, error accumulation during iterations may adversely affect robustness.

Although the above methods have demonstrated strong performance on the SEED-IV dataset, their modeling strategies differ fundamentally from unsupervised domain adaptation frameworks designed for cross-session transfer. Compared with existing graph-based domain adaptation methods, EGDA exhibits several key distinctions. Unlike CPTML, which tightly couples domain alignment with explicit graph-based metric learning in both source and target domains, EGDA adopts a lightweight subspace learning formulation with Laplacian regularization and avoids metric learning, resulting in improved computational efficiency and stability. In contrast to JAGP, EGDA decouples graph modeling from target label estimation and stabilizes conditional distribution alignment by incorporating within-class scatter information exclusively from the source domain. Moreover, unlike semi-supervised graph-based approaches such as GASDSL, EGDA explicitly aligns both marginal and class-conditional distributions, making it more suitable for unsupervised cross-session EEG emotion recognition.

\subsection{Transfer Learning Approaches}

\Gls{tca} is an unsupervised domain
adaptation method that learns a shared subspace by minimizing the discrepancy
between source and target marginal distributions using
\gls{mmd} \cite{pan2010domain}. By projecting data into transfer
components, TCA reduces domain shift at the distribution level. However, \gls{tca}
does not explicitly account for class-conditional distribution differences and
does not exploit source label information, which may limit its effectiveness
when conditional distributions vary significantly across domains.

\Gls{jda} extends \gls{tca} by explicitly aligning both
marginal and conditional distributions between source and target domains within
a unified subspace learning framework \cite{long2013transfer}. It leverages
pseudo-labels for the target domain to estimate class-conditional discrepancies
and iteratively refines the shared representation.

\Gls{vda} further addresses large domain shifts by jointly
reducing marginal and conditional distribution discrepancies while promoting
class discrimination through domain-invariant clustering \cite{tahmoresnezhad2017visual}.
By integrating distribution adaptation with within-class scatter minimization,
\gls{vda} learns a discriminative and transferable representation.

Collectively, these methods demonstrate that distribution alignment effectively reduces inter-session variability. Building on this foundation, our work integrates robust graph learning with
transfer alignment to better preserve discriminative structures during
cross-session adaptation.

\section{PROPOSED METHOD}
\subsection{Problem Statement}

Let $\mathcal{D}_s = \{\boldsymbol{X}_s, \boldsymbol{Y}_s\}$ denote a labeled
source-domain EEG dataset with $n_s$ samples, and
$\mathcal{D}_t = \{\boldsymbol{X}_t\}$ denote an unlabeled target-domain EEG
dataset with $n_t$ samples.
Here, $\boldsymbol{X}_s \in \mathbb{R}^{d \times n_s}$ and
$\boldsymbol{X}_t \in \mathbb{R}^{d \times n_t}$ represent the EEG feature matrices
of the source and target domains, respectively, where $d$ is the feature
dimensionality. We further define the combined feature matrix as
$
\boldsymbol{X} = [\boldsymbol{X}_s,\boldsymbol{X}_t] \in \mathbb{R}^{d \times n},
$
where $n = n_s + n_t$ is the total number of samples.
The source-domain label matrix is denoted by
$\boldsymbol{Y}_s \in \mathbb{R}^{C \times n_s}$, where $C$ is the number of emotion
classes (e.g., sadness, fear, happiness, and neutrality), and each column of
$\boldsymbol{Y}_s$ is a one-hot encoded label vector corresponding to a source
sample.

We assume that the source and target domains share the same feature space and
label space, i.e., $\mathcal{X}_s = \mathcal{X}_t$ and
$\mathcal{Y}_s = \mathcal{Y}_t$, but follow different data distributions due to
cross-session domain shift.
In particular, both the marginal feature distributions and the conditional label
distributions may differ, i.e., $P_s(x) \neq P_t(x)$ and
$P_s(y \mid x) \neq P_t(y \mid x)$.

To reduce the distribution discrepancy between domains, a linear projection
matrix $\boldsymbol{A} \in \mathbb{R}^{d \times d_f}$ is learned to map both source
and target EEG features into a shared subspace, yielding
$\boldsymbol{A}^T \boldsymbol{X}_s$ and $\boldsymbol{A}^T \boldsymbol{X}_t$.
This shared representation aims to align the source and target distributions
while preserving discriminative information for emotion recognition.

Under this setting, we address a multi-class EEG-based emotion recognition
problem without access to labeled target data.
The objective is to learn a transferable model that can accurately recognize
emotional states in the target domain despite cross-session distribution
mismatch.

\subsection{Model Formulation}
In this paper, we propose a joint adaptation framework that learns a low-dimensional representation $\boldsymbol{A} \in \mathbb{R}^{d \times d_f}$ and simultaneously improves the target-domain labels. Since the target domain is unlabeled (i.e., $P_t(\boldsymbol{Y}_t | \boldsymbol{X}_t)$ cannot be directly observed), we employ an \gls{em}-like strategy that iteratively updates both the projection matrix and the target pseudo-labels.

A central goal of transfer learning is to reduce distributional divergence between source and target domains. In \gls{eeg}-based emotion recognition, session variability, noise, and experimental conditions cause distribution shifts, which degrade the performance of classifiers trained solely on source data.
To address this, \gls{egda} learns a linear projection matrix $\boldsymbol{A}$ that maps both domains into a shared subspace where their marginal distributions are close.  
This objective can be formulated as

\begin{equation}\label{eq:MDA-minimization}
	\min_{\boldsymbol{A}} \left\| \frac{1}{n_s} \sum_{i=1}^{n_s} \boldsymbol{A}^T \boldsymbol{x}_{si} - \frac{1}{n_t} \sum_{j=1}^{n_t} \boldsymbol{A}^T \boldsymbol{x}_{tj} \right\|^2.
\end{equation}
This can be expressed in matrix form as:
\begin{equation}\label{eq:MDA-minimization-matrix}
	\min_{\boldsymbol{A}}\operatorname{Tr}(\boldsymbol{A}^T \boldsymbol{X} \boldsymbol{M}_0 \boldsymbol{X}^T \boldsymbol{A}),
\end{equation}
where $\boldsymbol{M}_0$ is \gls{mmd} coefficient matrix:
\begin{equation}\label{eq:M0}
	\left(\boldsymbol{M}_0\right)_{ij}= 
	\begin{cases}
		\frac{1}{n_s^2}, & \boldsymbol{x}_i, \boldsymbol{x}_j \in  \boldsymbol{X}_s, \\[1ex]
		\frac{1}{n_t^2}, & \boldsymbol{x}_i, \boldsymbol{x}_j \in \boldsymbol{X}_t, \\[1ex]
		\frac{-1}{n_s n_t}, & \text{otherwise}.
	\end{cases}
\end{equation}
Geometrically, this ensures that the Euclidean distance between the source and target domain centers in the latent space is minimized.

Aligning only marginal distributions does not guarantee semantic consistency across domains, since it ignores class information and can lead to samples from different classes being mapped close to each other in the latent space. Consequently, class-conditional misalignment may still exist. To address this issue, \gls{egda} aligns conditional distributions for each class using pseudo-labels for target samples.

Specifically, the class-wise conditional alignment is formulated as
\begin{equation}\label{eq:CDA-minimization}
	\min_{\boldsymbol{A}} \sum_{c=1}^C \left\| \frac{1}{n_s^{(c)}} \sum_{\boldsymbol{x}_i \in  \boldsymbol{X}_s^{(c)}} \boldsymbol{A}^T \boldsymbol{x}_i - \frac{1}{n_t^{(c)}} \sum_{\boldsymbol{x}_j \in \boldsymbol{X}_t^{(c)}} \boldsymbol{A}^T \boldsymbol{x}_j \right\|^2.
\end{equation}
$n_s^{(c)}$
and  $n_t^{(c)}$
denote the number of samples in the source and target domains that
belong to the class $c$, respectively. Also, $\boldsymbol{X}_s^{(c)}$
and $\boldsymbol{X}_t^{(c)}$
are defined to be the set of instances from class c belonging to the source and target data in turn.

This objective can be equivalently expressed in matrix form as
\begin{equation}\label{eq:CDA-minimization-matrix}
		\min_{\boldsymbol{A}} \operatorname{Tr}\!\left(\boldsymbol{A}^T \boldsymbol{X} \boldsymbol{M}_c \boldsymbol{X}^T \boldsymbol{A}\right).
\end{equation}
where, $\boldsymbol{M}_c \in \mathbb{R}^{n \times n}$ is \gls{mmd} coefficient matrix:
\begin{equation}\label{eq:Mc}
	(\boldsymbol{M}_c)_{ij} =
	\begin{cases}
		\frac{1}{(n_s^{{(c)}})^2} & \boldsymbol{x}_i, \boldsymbol{x}_j \in \boldsymbol{X}_s^{{(c)}} \\
		\frac{1}{(n_t^{(c)})^2} & \boldsymbol{x}_i, \boldsymbol{x}_j \in \boldsymbol{X}_t^{(c)} \\
		-\frac{1}{n_s^{(c)}n_t^{(c)}} & \boldsymbol{x}_i \in \boldsymbol{X}_s^{{(c)}},\ \boldsymbol{x}_j \in \boldsymbol{X}_t^{(c)} \\
		-\frac{1}{n_s^{{(c)}}n_t^{{(c)}}} & \boldsymbol{x}_j \in \boldsymbol{X}_s^{(c)},\ \boldsymbol{x}_i \in \boldsymbol{X}_t^{(c)} \\
		0 & \text{otherwise}
	\end{cases}
\end{equation}

To enhance separability in the latent space, samples belonging to the same class in the source domain should form compact clusters. This property can be encouraged by minimizing the within-class scatter matrix, defined as:
\begin{equation}\label{eq:Scatter_matrix}
	\boldsymbol{S}_w = \sum_{c=1}^C \sum_{\boldsymbol{x}_i \in \boldsymbol{X}_s^{(c)}} (\boldsymbol{x}_i - \boldsymbol{\mu}_c)(\boldsymbol{x}_i - \boldsymbol{\mu}_c)^T,
\end{equation}
where $\boldsymbol{\mu}_c$ is the mean vector of class $c$ in the source domain,
computed as:
\begin{equation}
	\boldsymbol{\mu}_c = \frac{1}{n_s^{(c)}} \sum_{\boldsymbol{x}_i \in \boldsymbol{X}_s^{(c)}} \boldsymbol{x}_i.
\end{equation}
and $n_s^{(c)}$ represents the number of source-domain samples belonging to class $c$.

Based on this formulation, \gls{egda} incorporates the following objective in the
projected space to promote intra-class compactness:
\begin{equation}\label{eq:within_class_objective}
	\min_{\boldsymbol{A}} \operatorname{Tr}\!\left(\boldsymbol{A}^T \boldsymbol{S}_w \boldsymbol{A}\right).
\end{equation}

To model local relationships among samples, we construct an adaptive similarity
graph represented by a weight matrix $\boldsymbol{S} \in \mathbb{R}^{n \times n}$,
where each element $s_{ij}$ quantifies the similarity between samples
$\boldsymbol{x}_i$ and $\boldsymbol{x}_j$.
Intuitively, samples that are closer in Euclidean distance are more likely to belong to the same class and are therefore assigned higher similarity weights, whereas more distant samples receive smaller weights to reflect weaker relationships.

The similarity learning problem is formulated as:
\begin{equation}\label{eq:Similarity_matrix_with_regularization}
	\begin{aligned}
		&\min _{\boldsymbol{S}} 
		\sum_{i,j=1}^n\Big(\|\boldsymbol{x}_i-\boldsymbol{x}_j\|_2^2 s_{ij}\Big),\\
		& \text{s.t.} \quad s_{ij} \geq 0, \; \boldsymbol{s}^i \boldsymbol{1}=1.
	\end{aligned}
	\end{equation}
where $\boldsymbol{s}^i$ denotes the $i$-th row of the similarity matrix $\boldsymbol{S}$,
and $\boldsymbol{1} \in \mathbb{R}^{n}$ is the all-ones vector.

However, problem \eqref{eq:Similarity_matrix_with_regularization} may lead to a trivial solution in which only one element of $\boldsymbol{s}^i$ equals one while all others are zero \cite{nie2014clustering}.
Although this minimizes the objective, it produces a degenerate graph 
where each sample is connected only to its nearest neighbor.
This fails to capture the local manifold structure of the data and consequently
undermines the effectiveness of graph-based regularization.

To address this, a regularization term is introduced:

\begin{equation}\label{eq:Similarity-Matrix}
	\begin{aligned}
		&\min _{\boldsymbol{S}} 
		\sum_{i,j=1}^n\Big(\|\boldsymbol{x}_i-\boldsymbol{x}_j\|_2^2 s_{ij}+\gamma s_{ij}^2\Big),\\
		& \text{s.t.} \quad s_{ij} \geq 0, \; \boldsymbol{s}^i \boldsymbol{1}=1.
	\end{aligned}
\end{equation}
Here, $\gamma$ is a control parameter that regulates the dispersion of values. A smaller $\gamma$ yields highly localized similarity weights, whereas a larger $\gamma$ produces more evenly distributed weights, improving stability in the presence of noise.

Since each row $\boldsymbol{s}^i$ depends only on distances involving sample $i$, 
problem~\eqref{eq:Similarity-Matrix} can be decomposed into $n$ independent 
row-wise subproblems:

\begin{equation}\label{eq:Similarity-Matrix_2}
		\begin{aligned}
	&\min _{\boldsymbol{S}} 
	\sum_{j=1}^n \Big(\|\boldsymbol{x}_i-\boldsymbol{x}_j\|_2^2 s_{ij}+\gamma s_{ij}^2\Big),\\
	& \text{s.t.} \quad s_{ij} \geq 0, \; \boldsymbol{s}^i \boldsymbol{1}=1.
		\end{aligned}
\end{equation}

Define
$m_{ij} = \|\boldsymbol{x}_i - \boldsymbol{x}_j\|_2^2$,
and let $\boldsymbol{m}^i \in \mathbb{R}^{1 \times n}$ denote the vector whose
$j$-th element is $m_{ij}$. Under this notation, problem~\eqref{eq:Similarity-Matrix}
can be rewritten in the following row-wise form:

\begin{equation}\label{eq:Opt}
	\begin{aligned}
	&\min _{\boldsymbol{s}^i} 
\Big\|\boldsymbol{s}^i + \frac{\boldsymbol{m}^i}{2\gamma}\Big\|_2^2,\\
		& \text{s.t.} \quad s_{ij} \geq 0, \; \boldsymbol{s}^i \boldsymbol{1}=1.
	\end{aligned}
\end{equation}
This reformulation indicates that $\boldsymbol{s}^i$ is obtained by projecting 
$-\frac{\boldsymbol{m}^i}{2\gamma}$ onto the probability simplex, which enforces 
non-negativity and unit-sum constraints on each row. As discussed in~\cite{nie2014clustering}, 
this projection yields the closest feasible similarity vector under these constraints.

Since \gls{eeg} data are often noisy and redundant, it is advantageous to learn the similarity matrix in a lower-dimensional subspace. Let 
$\boldsymbol{A} \in \mathbb{R}^{d \times d_f}$
be a linear projection matrix. The similarity learning problem is then extended as:

\begin{equation}\label{eq:projected-similarity-matrix}
	\begin{aligned}
		& \min _{\boldsymbol{S}, \boldsymbol{A}} 
		\sum_{i,j=1}^n \Big(\|\boldsymbol{A}^T\boldsymbol{x}_i-\boldsymbol{A}^T\boldsymbol{x}_j\|_2^2 s_{ij} + \gamma s_{ij}^2\Big), \\
		& \text{s.t.} \quad s_{ij} \geq 0, \; \boldsymbol{s}^i \boldsymbol{1}=1.
	\end{aligned}
\end{equation}

Once $\boldsymbol{S}$ is estimated, the resulting graph is used to compute the Laplacian matrix:
\begin{equation}\label{eq:laplacian-matrix}
	\boldsymbol{L}_s = \boldsymbol{D}_s - \frac{\boldsymbol{S} + \boldsymbol{S}^T}{2},
\end{equation}
where $\boldsymbol{D}_s$ is diagonal with entries $d_{ii}=\sum_j s_{ij}$.

The first term of \eqref{eq:projected-similarity-matrix} can be written in matrix form as \cite{peng2021gfil}

\begin{equation}
	\sum_{i,j=1}^n\|\boldsymbol{A}^T\boldsymbol{x}_i-\boldsymbol{A}^T\boldsymbol{x}_j\|_2^2 s_{ij}=\operatorname{Tr}\!\left(\boldsymbol{A}^T \boldsymbol{X}\boldsymbol{L}_s \boldsymbol{X}^T \boldsymbol{A}\right),
\end{equation}
while the second term can be expressed as:
\begin{equation}
	\sum_{i,j=1}^n s_{ij}^2 = \|\boldsymbol{S}\|_F^2 = \operatorname{Tr}(\boldsymbol{S}^T\boldsymbol{S}),
\end{equation}
Therefore, the matrix form of problem \eqref{eq:projected-similarity-matrix} is given by: 

\begin{equation}
	\min_{\boldsymbol{A},\boldsymbol{S}} \operatorname{Tr}\!\left(\boldsymbol{A}^T \boldsymbol{X} \boldsymbol{L}_s \boldsymbol{X}^T \boldsymbol{A} + \gamma \boldsymbol{S}^T \boldsymbol{S}\right).
\end{equation}

To prevent overfitting and improve generalization, Frobenius norm regularization
is applied to the projection matrix:

\begin{equation}
	\min_{\boldsymbol{A}} \|\boldsymbol{A}\|_F^2 = 	\min_{\boldsymbol{A}}  \operatorname{Tr}(\boldsymbol{A}^T \boldsymbol{A}).
\end{equation}
This penalizes large values in the projection matrix, encouraging smoother and more robust representations while reducing sensitivity
to noise in EEG data.

By integrating all components, the final optimization problem is formulated as:
\begin{equation}\label{eq:fina-Opt}
	\begin{aligned}
		&\min_{\boldsymbol{A},\,\boldsymbol{S}} 
		\alpha \sum_{c=0}^{C}\operatorname{Tr}\!\big(\boldsymbol{A}^T \boldsymbol{X}\boldsymbol{M}_c \boldsymbol{X}^T \boldsymbol{A}\big)
		+ \beta\,\operatorname{Tr}\!\big(\boldsymbol{A}^T \boldsymbol{S}_w \boldsymbol{A}\big)\\
		& + \mu\,\operatorname{Tr}\!\big(\boldsymbol{A}^T \boldsymbol{X}\boldsymbol{L}_s \boldsymbol{X}^T \boldsymbol{A}\big)
		+ \lambda\,\operatorname{Tr}\!\big(\boldsymbol{A}^T \boldsymbol{A}\big)
		+ \gamma\,\operatorname{Tr}\!\big(\boldsymbol{S}^T \boldsymbol{S}\big),\\[6pt]
		&\text{s.t.} \quad 
	 \boldsymbol{A}^T \boldsymbol{X}\boldsymbol{H}\boldsymbol{X}^T \boldsymbol{A} = \boldsymbol{I},\quad 
		\boldsymbol{S} \ge \boldsymbol{0},\quad 
		\boldsymbol{S}\boldsymbol{1} = \boldsymbol{1}.
	\end{aligned}
\end{equation}
where $\boldsymbol{H} = \boldsymbol{I} - \frac{1}{n}\boldsymbol{11}^T$ is the
centering matrix. Here, $c=0$ corresponds to marginal distribution alignment,
while $c=1,\dots,C$ denote class-conditional alignment terms. The parameters
$\alpha$, $\beta$, $\mu$, $\lambda$, and $\gamma$ are trade-off coefficients
controlling distribution alignment, within-class compactness, graph
regularization, projection regularization, and similarity smoothness,
respectively.


Due to the large search space, the hyperparameters were selected by randomly sampling from the predefined candidate set
$\{10^{-3}, 10^{-2}, 10^{-1}, 2\times10^{-1}, 5\times10^{-1},
10^{0}, 2\times10^{0}, 5\times10^{0}, 10^{1}, 5\times10^{1}, 10^{2}\}$.

\subsection{Model Optimization}

The objective function of \gls{egda} depends on two variables, the projection matrix $\boldsymbol{A}$ and the similarity matrix $\boldsymbol{S}$.
We optimize this objective using an alternating optimization strategy,
updating one variable while keeping the other fixed.

\textit{(1) $\boldsymbol{S}$ Optimization}:

To update $\boldsymbol{S}$ in problem~\eqref{eq:fina-Opt}, we optimize the projected similarity learning problem~\eqref{eq:projected-similarity-matrix} while keeping $\boldsymbol{A}$ fixed. 
Since each row $\boldsymbol{s}^i$ depends only on distances involving sample $i$,
problem \eqref{eq:projected-similarity-matrix} can be decomposed into $n$ independent 
row-wise subproblems as:

\begin{equation}\label{eq:projected-similarity-matrix_rowwise}
	\begin{aligned}
		& \min _{\boldsymbol{S}} 
		\sum_{j=1}^n \Big(\|\boldsymbol{A}^T\boldsymbol{x}_i-\boldsymbol{A}^T\boldsymbol{x}_j\|_2^2 s_{ij} + \gamma s_{ij}^2\Big), \\
		& \text{s.t.} \quad s_{ij} \geq 0, \; \boldsymbol{s}^i \boldsymbol{1}=1.
	\end{aligned}
\end{equation}
We denote
$
{d}_{ij}= {\|\boldsymbol{A}^T \boldsymbol{x}_i - \boldsymbol{A}^T \boldsymbol{x}_j\|_2^2},
$ 
and define $\boldsymbol{d}^i \in \mathbb{R}^{1 \times n}$ as the vector whose 
$j$-th element is $d_{ij}$. Problem~\eqref{eq:projected-similarity-matrix_rowwise}
can be rewritten in the following row-wise form:

\begin{equation}\label{eq:opt-solving-S}
			\begin{aligned}
		&\min _{\boldsymbol{s}^i}
  \|\boldsymbol{s}^i +  \frac{\boldsymbol{d}^i}{2\gamma}\|_2^2, \\
		& \text{s.t.} \quad s_{ij} \geq 0, \; \boldsymbol{s}^i \boldsymbol{1}=1.
	\end{aligned}
	\end{equation}
	where $\boldsymbol{s}^i$ denotes the $i$-th row of the similarity matrix $\boldsymbol{S}$, which is updated by solving \eqref{eq:opt-solving-S}. This update corresponds to a Euclidean projection onto the probability simplex and can be efficiently solved using the Lagrangian multiplier method combined with Newton’s method \cite{peng2021fuzzy}.

\textit{(2) $\boldsymbol{A}$ Optimization}:
 We can rewrite optimization problem \eqref{eq:fina-Opt} as:
\begin{equation}
	\begin{aligned}
	&	\min_{\boldsymbol{A}} \ 
\operatorname{Tr}\!\left(
\boldsymbol{A}^T
\left(
\begin{aligned}
	&\alpha \sum_{c=0}^{C} \boldsymbol{X}\boldsymbol{M}_c\boldsymbol{X}^T \\
	&+ \beta \boldsymbol{S}_w
	+ \mu \boldsymbol{X}\boldsymbol{L}_s\boldsymbol{X}^T
	+ \lambda \boldsymbol{I}
\end{aligned}
\right)
\boldsymbol{A}
\right), \\
&	\text{s.t. } \quad \boldsymbol{A}^T \boldsymbol{X} \boldsymbol{H} \boldsymbol{X}^T \boldsymbol{A} =\boldsymbol{I}.
\end{aligned}
\end{equation}

To update $\boldsymbol{A}$, we employ the method of Lagrange multipliers. Accordingly, the corresponding Lagrangian function is defined as follows:

\begin{equation}\label{eq:lagrang of Opt}
	\begin{aligned}
	&	\mathcal{L}(\boldsymbol{A}) =\\
		&\,\operatorname{Tr}\!\left(
		\boldsymbol{A}^T
		(
	\alpha \sum_{c=0}^{C}\boldsymbol{X}\boldsymbol{M}_c \boldsymbol{X}^T
		+ \mu \boldsymbol{X}\boldsymbol{L}_s \boldsymbol{X}^T
		+ \beta \boldsymbol{S}_w
		+ \lambda \boldsymbol{I}
		)
		\boldsymbol{A}
		\right) \\
		&\,+ \operatorname{Tr}\!\left(
		\boldsymbol{\Phi}
		(
		\boldsymbol{I}
		- \boldsymbol{A}^T \boldsymbol{X}\boldsymbol{H}\boldsymbol{X}^T \boldsymbol{A}
		)
		\right).
	\end{aligned}
\end{equation}
where $\boldsymbol{\Phi} \in \mathbb{R}^{d_f \times d_f}$ is a diagonal matrix of Lagrange multipliers~\cite{ghojogh1903eigenvalue}.

Taking the derivative of the Lagrangian with respect to $\boldsymbol{A}$ and setting it  to zero, the following equation is obtained:

\begin{equation}\label{eq:eigendecomposition}
		\begin{aligned}
&	\left(\alpha \sum_{c=0}^{C} \boldsymbol{X}   \boldsymbol{M}_c  \boldsymbol{X}^T  + \beta \boldsymbol{S}_w + \mu \boldsymbol{X} \boldsymbol{L}_s \boldsymbol{X}^T + \lambda \boldsymbol{I}\right)\boldsymbol{A} \\
 &	=\boldsymbol{X} \boldsymbol{H} \boldsymbol{X}^T \boldsymbol{A}\boldsymbol{\Phi}.
		\end{aligned}
\end{equation}

Let
\begin{equation}
	\begin{aligned}
		\boldsymbol{P} &= \alpha \sum_{c=0}^{C} \boldsymbol{X} \boldsymbol{M}_c \boldsymbol{X}^T
		+ \beta \boldsymbol{S}_w
		+ \mu \boldsymbol{X}\boldsymbol{L}_s \boldsymbol{X}^T
		+ \lambda \boldsymbol{I},\\
		\boldsymbol{Q} &= \boldsymbol{X}\boldsymbol{H}\boldsymbol{X}^T.
	\end{aligned}
\end{equation}
For notational convenience, the generalized eigenvalue problem can be written as:
\begin{equation}\label{eq:simpleofeigen}
	\boldsymbol{P}\boldsymbol{A} = \boldsymbol{Q}\boldsymbol{A}\boldsymbol{\Phi}.
\end{equation}

The columns of $\boldsymbol{A}$ correspond to the generalized eigenvectors,
and the diagonal entries of $\boldsymbol{\Phi}$ are the associated eigenvalues.
 To minimize the objective, we select the $d_f$ eigenvectors associated with the smallest eigenvalues. Equivalently, $\boldsymbol{A}$ can be obtained from the $d_f$ eigenvectors of $\boldsymbol{Q}^{-1}\boldsymbol{P}$ corresponding to the smallest eigenvalues, assuming that the null space of $\boldsymbol{X}$ has been removed, such that
 $\boldsymbol{Q}$ becomes invertible.

After projecting the data into the learned subspace $\boldsymbol{A}$, a 1-nearest neighbor (1-NN) classifier is employed to assign pseudo-labels to the target samples.
These pseudo-labels are then used to update $\boldsymbol{M}_c$ and \( \boldsymbol{S} \), refining class-wise distribution alignment. This procedure is iteratively performed for a fixed number of iterations, which is empirically determined. At each iteration, \gls{egda} integrates pseudo-label updating with an \gls{em}-like optimization strategy, progressively improving the target label estimates.

Based on the above analysis, the whole procedure of \gls{egda}
is summarized in Algorithm \ref{alg:algorithmofEDGA}.

\begin{algorithm}
	\caption{EEG-based Graph Domain Adaptation for Robust Cross-Session EEG Emotion Recognition (\gls{egda}).}\label{alg:algorithmofEDGA}
	\begin{algorithmic}
		\STATE \textbf{Input:} \( \{\boldsymbol{X}_s,\boldsymbol{Y}_s\} \), target \( \boldsymbol{X}_t \); parameters \( \alpha,\beta,\mu,\lambda,\gamma \); subspace dimension \( d_f \).
		\STATE \textbf{Output:} Predicted labels of the target samples $\boldsymbol{Y}_t$ and projection matrix $\boldsymbol{A}$;
		\STATE Initialize  \( \boldsymbol{S} \) by solving  \eqref{eq:Similarity-Matrix}
		\STATE 	Calculate the Laplacian matrix according to \eqref{eq:laplacian-matrix};
		\STATE Calculate the scatter matrix according to \eqref{eq:Scatter_matrix};
		
		\STATE 	Calculate the $\boldsymbol{M}_0$ according to \eqref{eq:M0} and the $\boldsymbol{M}_c$ according to \eqref{eq:Mc}.
	
			\REPEAT

		\STATE Update $\boldsymbol{A}$ by solving the eigendecomposition in~\eqref{eq:eigendecomposition}  and selecting the $d_f$ eigenvectors corresponding to the smallest eigenvalues.
		
		\STATE Train a classifier \( f \) on \( (\boldsymbol{A}^T\boldsymbol{X}_s,\boldsymbol{Y}_s) \);	\\
		\STATE  Update pseudo-labels \( \hat{\boldsymbol{Y}}_t=f(\boldsymbol{A}^T\boldsymbol{X}_t) \);\\
		\STATE Update \( \boldsymbol{M}_c \) using current pseudo-labels according to \eqref{eq:Mc};\\
		\STATE Update \( \boldsymbol{S} \) according to \eqref{eq:projected-similarity-matrix};\\
			\STATE 	  Recompute the Laplacian matrix according to \eqref{eq:laplacian-matrix}.\\
\UNTIL{the maximum number of iterations is reached}

\end{algorithmic}
\end{algorithm}

\section{Experiments}

In this section, we evaluate the performance of the proposed \gls{egda} model. We first describe the \gls{eeg} dataset and the preprocessing steps. Then, we present the experimental setup and conduct two types of comparisons: (1) comparisons with existing methods evaluated on the same dataset, and (2) comparisons with representative methods that share similar modeling strategies. Finally, we introduce the performance measures used for evaluation. All experiments are conducted under a subject-dependent, cross-session setting,
where EEG data from the same subject but different sessions are used as source
and target domains, respectively.

\subsection{Dataset Description}

In this study, we conduct experiments on the publicly available SEED-IV dataset \cite{zheng2018emotionmeter}, which is widely used for EEG-based emotion recognition. The dataset consists of EEG recordings from 15 healthy subjects, each participating in three recording sessions conducted on different days, resulting in a total of 45 sessions. In each session, subjects viewed 24 carefully selected video clips designed to elicit four emotional states: sadness, fear, happiness, and neutrality, with six clips corresponding to each emotion.

EEG signals were acquired using 62 electrodes arranged according to the international 10--20 system. The raw signals were preprocessed and segmented into 4-second EEG samples, which serve as the basic units for feature extraction. \gls{de} features were then extracted under the assumption that EEG signals within each frequency band approximately follow a Gaussian distribution. DE features were computed from five standard frequency bands, namely Delta, Theta, Alpha, Beta, and Gamma. By concatenating DE features across all channels and frequency bands, each EEG sample is represented by a 310-dimensional feature vector. Owing to differences in video clip durations, each session contains approximately 830 EEG samples.

%
\subsection{Overall Cross-Session Performance}

The proposed \gls{egda} framework was evaluated on the SEED-IV dataset under a cross-session domain adaptation setting. EEG data from two different sessions of the same subject were used to construct each transfer task, where one session served as the labeled source domain and the other as the unlabeled target domain. Three cross-session transfer tasks were considered: Session~1$\rightarrow$2, Session~1$\rightarrow$3, and Session~2$\rightarrow$3.

As shown in Tables~\ref{tab:datasetbased-results12}--\ref{tab:datasetbased-results23}, \gls{egda} consistently outperforms most existing methods whose performance has been previously reported on the SEED-IV dataset including GASDSL, OGSSL, SRAGL, GFIL, JSFA, CPTML and JAGP, achieving average accuracies of 81.22\%, 80.15\%, and 83.27\% for the three transfer tasks, respectively. Although CPTML attains slightly higher accuracy, it involves higher computational complexity, which may limit its applicability
in large-scale or real-time scenarios.
Overall, \gls{egda}  achieves the second-highest average accuracy among the compared methods.

When compared with representative transfer learning methods (JDA, SDA, and VDA; Tables~\ref{tab:transfer-leraning-result12}--\ref{tab:transfer-leraning-result23}), \gls{egda} demonstrates improved recognition performance across all evaluated tasks. The results indicate that \gls{egda} provides consistent performance gains over JDA and SDA, while exhibiting competitive performance relative to VDA. These observations suggest that \gls{egda} is effective in preserving neighborhood structure during the domain adaptation process.

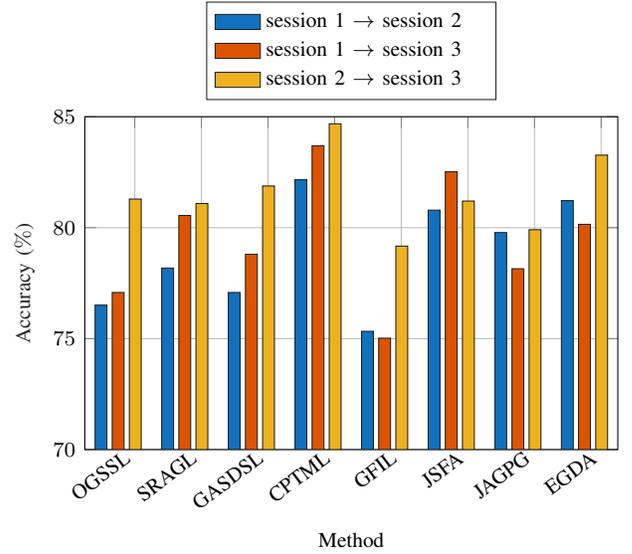
\begin{figure}
	\centering
	\setlength\axheight{0.5\linewidth}%
	\setlength\axwidth{0.8\linewidth}%
	\setlength\myfontsize{8pt}%
	\definecolor{mycolor1}{rgb}{0.06600,0.44300,0.74500}%
\definecolor{mycolor2}{rgb}{0.12941,0.12941,0.12941}%
\definecolor{mycolor3}{rgb}{0.86600,0.32900,0.00000}%
\definecolor{mycolor4}{rgb}{0.92900,0.69400,0.12500}%
\begin{tikzpicture}[font=\fontsize{\myfontsize}{1.2\myfontsize}\selectfont,>=latex]
\begin{axis}[%
width=\axwidth,
height=\axheight,
at={(0in,0in)},
scale only axis,
bar shift auto,
xmin=0.5,
xmax=8.5,
xtick={1.25,2.25,3.25,4.25,5.25,6.25,7.25,8.25},
xticklabels={{OGSSL},{SRAGL},{GASDSL},{CPTML},{GFIL},{JSFA},{JAGPG},{EGDA}},
xticklabel style={font=\footnotesize, rotate=37, anchor=east},
xlabel={Method},
ymin=70,
ymax=85,
ylabel style={font=\fontsize{\myfontsize}{1.2\myfontsize}\selectfont, color=white!15!black},
ylabel={Accuracy ($\%$)},
axis background/.style={fill=white},
xmajorgrids,
ymajorgrids,
legend style={at={(0.5,1.05)}, anchor=south, legend cell align=left, align=left, draw=white!15!black, legend columns=1}
]
\addplot[ybar, bar width=0.178, fill=mycolor1, draw=mycolor2, area legend] table[row sep=crcr] {%
1	76.51\\
2	78.18\\
3	77.08\\
4	82.16\\
5	75.33\\
6	80.79\\
7	79.78\\
8	81.22\\
};
\addplot[forget plot, color=mycolor2] table[row sep=crcr] {%
0.511111111111111	0\\
8.48888888888889	0\\
};
\addlegendentry{session 1 $\rightarrow$ session 2\quad}

\addplot[ybar, bar width=0.178, fill=mycolor3, draw=mycolor2, area legend] table[row sep=crcr] {%
1	77.08\\
2	80.55\\
3	78.8\\
4	83.69\\
5	75.02\\
6	82.52\\
7	78.15\\
8	80.15\\
};
\addplot[forget plot, color=mycolor2] table[row sep=crcr] {%
0.5	0\\
8.5	0\\
};
\addlegendentry{session 1 $\rightarrow$ session 3\quad}

\addplot[ybar, bar width=0.178, fill=mycolor4, draw=mycolor2, area legend] table[row sep=crcr] {%
1	81.29\\
2	81.09\\
3	81.88\\
4	84.68\\
5	79.17\\
6	81.2\\
7	79.91\\
8	83.27\\
};
\addplot[forget plot, color=mycolor2] table[row sep=crcr] {%
0.511111111111111	0\\
8.48888888888889	0\\
};
\addlegendentry{session 2 $\rightarrow$ session 3}

\end{axis}
\end{tikzpicture}%
	\caption{Comparison of recognition accuracies for different models in cross-session EEG emotion recognition}
	\label{fig:compare accuracy of models}
\end{figure}
\subsection{Emotion-Specific Recognition}
\begin{figure}
	\centering
	\setlength\myfontsize{8pt}%
	\begin{tikzpicture}[font=\fontsize{\myfontsize}{1.2\myfontsize}\selectfont,>=latex]
	\node (n) [fill = purple!20,draw = purple, thick, circle, minimum size = 1.25cm] at (-2,2) {Neutral};
	\node (h) [fill = orange!20, draw = orange, thick, circle, minimum size = 1.25cm] at (2,2) {Happy};
	\node (f) [fill = blue!20, draw = blue, thick, circle, minimum size = 1.25cm] at (2,-2) {Fear};
	\node (s) [fill = ForestGreen!20, draw = ForestGreen, thick, circle, minimum size = 1.25cm] at (-2,-2) {Sad};
	\draw [thick,purple,->] (n) edge[bend right=15 ] node [fill = white] {$7.91$} (h);
	\draw [thick,purple,->] (n) edge[bend right=15 ] node [fill = white, rotate = 90] {$5.74$} (s);
	\draw [thick,purple,->] (n) edge[bend right=15 ] node [fill = white, rotate = -45] {$8.82$} (f);
	\draw[thick,purple,->] (n) edge[out=105, in=165, looseness=4] node [above,rotate = 45] {$87.86$} (n);
	\draw [ForestGreen, thick, ->] (s) edge[bend right=15 ] node [fill = white, rotate = -90] {$9.95$} (n);
	\draw [ForestGreen, thick, ->] (s) edge[bend right=15 ] node [fill = white, rotate = 45] {$7.22$} (h);
	\draw [ForestGreen, thick, ->] (s) edge[bend right=15 ] node [fill = white] {$10.84$} (f);
	\draw[ForestGreen, thick, ->] (s) edge[out=195, in=255, looseness=4] node [below,rotate = -45] {$70.19$} (s);
	\draw [thick , orange, ->] (h) edge[bend right=15 ] node [fill = white] {$3.24$} (n);
	\draw [thick , orange, ->] (h) edge[bend right=15 ] node [fill = white, rotate = 45] {$5.74$} (s);
	\draw [thick , orange, ->] (h) edge[bend right=15 ] node [fill = white, rotate = 90] {$8.35$} (f);
	\draw[thick , orange, ->] (h) edge[out=15, in=75, looseness=4] node [above,rotate = -45] {$76.52$} (h);
	
	\draw [blue, thick, ->] (f) edge[bend right=15 ] node [fill = white, rotate = -45] {$3.15$} (n);
	\draw [blue, thick, ->] (f) edge[bend right=15 ] node [fill = white] {$14.39$} (s);
	\draw [blue, thick, ->] (f) edge[bend right=15 ] node [fill = white, rotate = -90] {$4.44$} (h);
	\draw[blue, thick, ->] (f) edge[out=285, in=345, looseness=4] node [below,rotate = 45] {$75.90$} (f);
\end{tikzpicture}
	\caption{The average confusion matrix of \gls{egda} for cross-session EEG emotion recognition. }
	\label{fig:confusion_matrix}
\end{figure}
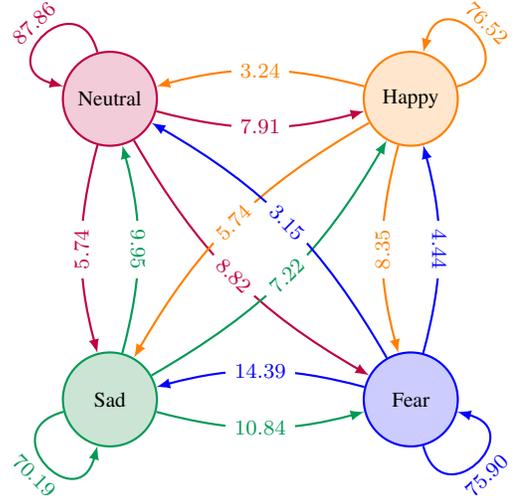

To further investigate class-wise recognition performance, Fig.~\ref{fig:confusion_matrix} presents the averaged confusion matrix over three transfer tasks. The results show that the neutral emotional state is recognized with the highest accuracy (87.83\%), followed by happiness (76.52\%), whereas fear and sadness are more prone to misclassification.

\begin{table*}[htbp]
	\centering
	
	\caption{Cross-session (session 1 $\rightarrow$ session 2) emotion recognition results (\%) of the eight models on SEED\_IV dataset.}
	\label{tab:datasetbased-results12}
		\begin{tabular}{c c c c c c c c c}
			\hline
			subject & GASDSL & OGSSL & SRAGL & CPTML & GFIL & JSFA & JAGP & \gls{egda} \\
			\hline
			1  &80.17 & 65.14 & 65.26 &\textbf{88.70} &66.59&75.00&76.56& 80.17\\
			2  & 89.06 & 97.36 & 88.94 & 97.36&85.70&97.36&95.55 &\textbf{98.80}  \\
			3  & 77.40 & 75.48 & 79.81 & \textbf{91.23} &67.79& 84.98 &74.64&88.10\\
			4  & 76.32 & 65.99 & 71.27 & 85.94& 84.98&  82.93 &68.39&\textbf{88.58}\\
			5  & 69.35 & 69.95 &68.99  & \textbf{88.58}&75.12&80.65 &86.06&81.61\\
			6  & 77.16 & 71.39 & 81.37 & 66.95&73.32&74.76& \textbf{84.13}&70.43\\
			7  & 79.33 & 85.94 &82.93  & 94.35 &94.35&\textbf{95.55}&93.15&89.66\\
			8  & 80.89 & 83.05 &81.13  & 83.97 &78.97&81.85&89.30&\textbf{84.13}\\
			9  & 80.77 & 78.73 & 82.21 & 73.80&79.33&\textbf{82.69}& 82.09&75.24\\
			10 & 72.48 & 74.88 & 73.44 & \textbf{75.24}&59.25&70.19& 64.54&72.12\\
			11 & 68.75 & 67.79 & 69.59 & 64.54&\textbf{69.59}&66.83&64.30&59.25\\
			12 & 60.58 & 53.25 & 74.16 & 66.11&70.07&\textbf{74.76}&69.83&72.00\\
			13 & 69.79 & \textbf{75.60} & 72.60 & 72.84&70.67&69.83&68.39&68.63 \\
			14 & 79.63 & 85.70 & 83.65 & 84.01&82.21&78.73&81.73&\textbf{89.66} \\
			15 & 94.47 & 97.36 & 97.36 & 98.80&93.15&95.67&\textbf{100}& \textbf{100}\\
			\hline
			Average & 77.08 & 76.51 & 78.18 & \textbf{82.16} &75.33&80.79&79.91&81.22\\
			\hline
		\end{tabular}
\end{table*}

\begin{table*}[htbp]
	\centering 
	\caption{Cross-session (session 1 $\rightarrow$ session 3) emotion recognition results (\%) of the eight models on SEED\_IV dataset.}
	\label{tab:datasetbased-results13}
		\begin{tabular}{c c c c c c c c c}
			\hline
			subject & GASDSL & OGSSL & SRAGL & CPTML & GFIL & JSFA & JAGP & \gls{egda} \\
			\hline
			1  & 75.84 & 80.54 & 73.48 & \textbf{84.67} &74.09&87.47&79.44&\textbf{84.67}\\
			2  & 87.23 & 91.85 &\textbf{93.43} & 88.69&80.54&80.90&68.61 &57.18  \\
			3  & \textbf{83.45} & 66.42 & 67.03 & 74.21 &75.06& 80.41 &71.05&65.09\\
			4  & 79.83 & 79.68 &80.05 & \textbf{93.07}& 86.74&  82.60 &80.78&83.21\\
			5  & 73.84 &52.43 &65.21  & \textbf{85.40}&66.18&80.17 &76.52&80.66\\
			6  & 76.64 & 85.89 & 81.39 & 84.79&82.12&\textbf{94.89}& 88.81&91.36\\
			7  & 88.81 &91.85 &96.35 &90.15 &79.20&\textbf{95.01}&86.74&89.17\\
			8  & 79.56 &80.90 & 85.04  & \textbf{94.89} &81.02&94.04&92.82&92.94\\
			9  & 75.43 & 77.98 &79.56& \textbf{80.54}&70.19&77.25& 75.91&\textbf{80.54}\\
			10 & 73.48 & 73.97 & 75.30 &73.51&68.61&75.43& 69.34&\textbf{76.28}\\
			11 & 78.59 &78.83 & 80.66 & \textbf{85.89}&71.53&75.18&72.51&78.83\\
			12 & 65.94 &55.84 & 74.94 & \textbf{74.45}&56.57&64.60&58.34&66.30\\
			13 & \textbf{74.09} & 64.48 & 73.11 & 63.02&57.91&64.72&64.72&62.41 \\
			14 & 83.73 &84.31 & 89.66 & 89.54&85.28&92.58&89.42&\textbf{97.69} \\
			15 & 85.52 & 91.24 &93.07 &92.58&90.27&92.58&\textbf{97.32}& 95.86\\
			\hline
			Average & 78.80 & 77.08 & 80.55 & \textbf{83.69} &75.02&82.52&78.15&80.15\\
			\hline
		\end{tabular}
\end{table*}

\begin{table*}[htbp]
	\centering
	\caption{Cross-session (session 2 $\rightarrow$ session 3) emotion recognition results (\%) of the eight models on SEED\_IV dataset.}
	\label{tab:datasetbased-results23}
		\begin{tabular}{c c c c c c c c c}
			\hline
			subject & GASDSL & OGSSL & SRAGL & CPTML & GFIL & JSFA &JAGP & \gls{egda} \\
			\hline
	     1  & 75.30 & 73.36 & 72.63 & 76.16 & 65.69 & 69.83 &  72.75   &\textbf{81.63} \\
	     
	      2  & 86.37 & \textbf{90.63} & 88.81 & 74.21 & 82.60 & 64.84 &  65.82 &68.13 \\
	     
		 3  & 82.73 & 79.20 & 76.64 & \textbf{86.98} & 70.92 & 78.47 & 80.66&77.49 \\
		 4  & 84.43 & 82.73 & 83.45 & \textbf{97.93} & 90.15 & 88.69 &89.90&92.94 \\
		 5  & 77.01 & 77.98 & 83.21 & 83.82 & 72.26 &\textbf{85.40} & 78.83 &68.61 \\
		 6  & 85.04 & 89.90 & 92.70 & 89.42 & 92.34 & \textbf{94.28} &84.31 &89.42\\
		 7  & 92.94 & 90.15 & 90.02 & 94.89 & 89.78 & \textbf{96.84} &  90.15   &91.85 \\
	    8  & 81.75 & 84.18 & 84.06 & \textbf{88.93} & 74.70 & 77.13 &  80.78   &85.89 \\
		 9  & 75.43 &\textbf{78.59} & 77.25 & 72.51 & 75.67 & 75.55 &  78.35   &73.24 \\
		10 & 85.52 & 85.16 & 78.59 & 89.42 & 81.14 & 77.37 &  79.32   &\textbf{90.51} \\
		 11 & 78.59 & 61.19 & 70.68 & 66.91 & 64.72 & \textbf{83.94} &   65.94  &71.17 \\
		 12 & 68.98 & 66.79 & 77.86 & \textbf{82.41} & 79.81 & 75.67 &  69.22   &80.41 \\
		 13 & 76.16 & 76.40 & 72.63 & 75.91 & 61.68 & 63.75 &  73.97   &\textbf{80.41} \\
	 14 & 90.75 & 92.70 & 91.00 & 95.26 & 89.17 & 91.73 &   92.82  &\textbf{100.00} \\
		 15 & 87.23 & 90.39 & 90.27 & 95.38 & 93.07 & 94.53 &   93.92  &\textbf{97.32} \\
			\hline
			Average   & 81.88 & 81.29 & 81.99 & \textbf{84.68} & 79.17 & 81.20 &79.78& 83.27 \\
			\hline
		\end{tabular}
\end{table*}

\begin{table}[htbp]
	\centering	\caption{Cross-session (session 1 $\rightarrow$ session 2) emotion recognition results (\%) of the four models on SEED\_IV dataset.}
	\label{tab:transfer-leraning-result12}
		\begin{tabular}{c c c c c c}
			\hline
		Subject & JDA & SDA & VDA & EGDA \\
			\hline
			1  & 63.22 & 70.79 & 80.17 &  \textbf{80.17} \\
			2  & 92.55 & 91.23 & 97.36 &  \textbf{98.80} \\
			3  & 68.99 & 69.59 & 84.50 & \textbf{88.10} \\
			4  & 64.30 & 81.37 & 86.06 &  \textbf{88.58} \\
			5  & 71.15 & 59.98 & 70.07 & \textbf{81.61} \\
			6  & 64.98 & 63.22 & 57.81 & \textbf{70.43}  \\
			7  & 73.08 & 85.10 & 92.43 &  \textbf{89.66} \\
			8  & 74.04 & 76.20 & 83.41 & \textbf{84.13} \\
			9  & 72.64 & 59.01 & 70.19 & \textbf{75.24}  \\
			10 & 64.18 & 53.25 & 69.11 & \textbf{72.12}  \\
			11 & \textbf{62.98} & 63.58 & 53.12 & 59.25 \\
			12 & 55.41 & 60.94 & 71.15 & \textbf{72.00} \\
			13 & 64.42 & 61.21 & 66.95 & \textbf{68.63}  \\
			14 & 67.79 & 77.76 & 84.86 & \textbf{89.66} \\
			15 & 93.03 & 95.67 & \textbf{100} & \textbf{100}  \\
			\hline
			Average & 70.18 & 71.26 & 76.21 & \textbf{81.22}  \\
			\hline
		\end{tabular}
\end{table}
\begin{table}[htbp]
	\centering
\caption{Cross-session (session 1 $\rightarrow$ session 3) emotion recognition results (\%) of the four models on SEED\_IV dataset.}
	\label{tab:transfer-leraning-result13}
	\begin{tabular}{c c c c c c}
			\hline
			Subject & JDA & SDA & VDA & EGDA \\
			\hline
			1  & 62.53 & 77.01 & 83.70  & \textbf{84.67} \\
			2  & \textbf{65.33} & 58.52 & 55.47 & 57.18 \\
			3  & 53.28 & 42.09 & 68.13 &  \textbf{65.09} \\
			4  & 77.74 & \textbf{86.25} & 83.33 & 83.21 \\
			5  & 70.19 & 72.51 & 78.47 &  \textbf{80.66} \\
			6  & 80.90 & 75.79 & 82.85 & \textbf{91.36}  \\
			7  & 56.57 & 85.16 & 87.47 &  \textbf{89.17} \\
			8  & 81.75 & 89.42 & 86.50 & \textbf{92.94} \\
			9  & 58.27 & 52.43 & \textbf{81.02} &  80.54 \\
			10 & 64.84 & 70.21 & 73.24 &  \textbf{76.28} \\
			11 & 66.06 & 67.82 & 68.25 & \textbf{78.83} \\
			12 & 54.26 & 51.70 & 58.15 &  \textbf{66.30} \\
			13 & 56.84 & 40.88 &\textbf{ 65.08} & 62.41  \\
			14 & 83.82 & 81.39 & 93.92 & \textbf{97.69} \\
			15 & 80.54 & 78.22 & 92.21 & \textbf{95.86}    \\
			\hline
			Average & 67.53 & 68.63 & 75.17 & \textbf{80.15}\\
			\hline
		\end{tabular}
\end{table}
\begin{table}[htbp]
	\centering
\caption{Cross-session (session 2 $\rightarrow$ session 3) emotion recognition results (\%) of the four models on SEED\_IV dataset.}
	\label{tab:transfer-leraning-result23}
		\begin{tabular}{c c c c c c}
			\hline
			Subject & JDA & SDA & VDA & EGDA \\
			\hline
			1  & 58.88 & 66.30 & 65.45 &  \textbf{81.63} \\
			2  & 59.61 & 64.72 & 61.19 & \textbf{68.13} \\
			3  & 84.55 & 66.67 & 72.38 & \textbf{77.49} \\
			4  & 67.40 & 85.04 & 90.15 & \textbf{92.94} \\
			5  & 72.63 & 71.29 & \textbf{72.38} & 68.61 \\
			6  & 55.96 & 86.42 & 87.35 &  \textbf{89.42} \\
			7  & 90.15 & 91.73 & 90.02 &  \textbf{91.85} \\
			8  & 65.69 & 83.82 & 79.20 & \textbf{85.89} \\
			9  & 66.79 & 51.46 & 64.36 & \textbf{73.24}  \\
			10 & 85.40 & 82.48 & 86.01 &  \textbf{90.51} \\
			11 & 59.37 & 62.04 & \textbf{71.17} & \textbf{71.17}  \\
			12 & 51.82 & 78.47 &\textbf{80.41} & \textbf{80.41} \\
			13 & 66.91 & 67.72 & \textbf{80.41} & \textbf{80.41} \\
			14 & 79.68 & 87.71 & \textbf{100} & \textbf{100} \\
			15 & 90.39 & 90.39 & 92.70& \textbf{97.32} \\
			\hline
			Average & 70.35 & 75.75 & 77.64 & \textbf{83.27 } \\
			\hline
		\end{tabular}
\end{table}


\subsection{ Critical Frequency Bands and Channels Identification}

In this study, EEG feature vectors are constructed by concatenating \gls{de}
features extracted from multiple frequency bands and channels.
To evaluate the discriminative contribution of individual EEG features, we adopt the normalized $\ell_{2}$-norm of each row of the learned projection matrix $\boldsymbol{A}$ as a feature importance measure. Specifically, the importance of the $i$-th EEG feature is defined as:
\begin{equation}
	{\boldsymbol{\theta}}_{i} = \frac{\| \boldsymbol{a}^i \|_{2}}{\sum_{j=1}^{d} \| \boldsymbol{a}^j \|_{2}}, \quad i = 1, 2, \ldots, d, \quad \boldsymbol{\theta} \in \mathbb{R}^{d},
	\label{eq:feature_importance}
\end{equation}
where larger values of $\boldsymbol{\theta}_{i}$ indicate stronger discriminative capability for distinguishing emotional states.

\begin{figure}\centering
	\setlength\axheight{0.5\linewidth}%
	\setlength\axwidth{0.8\linewidth}%
	\setlength\myfontsize{8pt}%
		\begin{tikzpicture}[every text node part/.style={align=center},>=latex,node distance=1cm,decoration={snake, amplitude=1.5pt, segment length=0.5cm}]
	\fill [red!20, rounded corners=2mm] (0.1,-2.5) rectangle (0.9,2.5);
	
;
	\node [Brown] at (0.5,2) {$\theta_1$};
	\node [blue] at (0.5,1) {$\theta_2$};
	\draw [fill] (0.5,-0.5) circle (1pt);
	\draw [fill] (0.5,-0.25) circle (1pt);
	\draw [fill] (0.5,-0.75) circle (1pt);
	\node [orange] at (0.5,-2) {$\theta_{62}$};
	
	\begin{scope}[xshift = 1cm]
		\fill [yellow!20,rounded corners=2mm] (0.1,-2.5) rectangle (0.9,2.5);
		\node [Brown] at (0.5,2) {$\theta_{63}$};
		\node [blue] at (0.5,1) {$\theta_{64}$};
		\draw [fill] (0.5,-0.5) circle (1pt);
		\draw [fill] (0.5,-0.25) circle (1pt);
		\draw [fill] (0.5,-0.75) circle (1pt);
		\node [orange] at (0.5,-2) {$\theta_{124}$};
	\end{scope}
	\begin{scope}[xshift = 2cm]
 	\fill [cyan!20,rounded corners=2mm] (0.1,-2.5) rectangle (0.9,2.5);
		\node [Brown] at (0.5,2) {$\theta_{125}$};
		\node [blue] at (0.5,1) {$\theta_{126}$};
		\draw [fill] (0.5,-0.5) circle (1pt);
		\draw [fill] (0.5,-0.25) circle (1pt);
		\draw [fill] (0.5,-0.75) circle (1pt);
		\node [orange] at (0.5,-2) {$\theta_{186}$};
	\end{scope}

	\begin{scope}[xshift = 3cm]
		\fill [green!20,rounded corners=2mm] (0.1,-2.5) rectangle (0.9,2.5);
		\node [Brown] at (0.5,2) {$\theta_{187}$};
		\node [blue] at (0.5,1) {$\theta_{188}$};
		\draw [fill] (0.5,-0.5) circle (1pt);
		\draw [fill] (0.5,-0.25) circle (1pt);
		\draw [fill] (0.5,-0.75) circle (1pt);
		\node [orange] at (0.5,-2) {$\theta_{248}$};
	\end{scope}
	
	\begin{scope}[xshift = 4cm]
		\fill [Purple!20,rounded corners=2mm] (0.1,-2.5) rectangle (0.9,2.5);
		\node [Brown] at (0.5,2) {$\theta_{249}$};
		\node [blue] at (0.5,1) {$\theta_{250}$};
		\draw [fill] (0.5,-0.5) circle (1pt);
		\draw [fill] (0.5,-0.25) circle (1pt);
		\draw [fill] (0.5,-0.75) circle (1pt);
		\node [orange] at (0.5,-2) {$\theta_{310}$};
	\end{scope}
	\draw [Brown, line width = 1pt] (6,2) circle (0.4cm);
	\node [Brown,inner sep = 0] at (6,2) {FP1};
	\draw [blue, line width = 1pt] (6,1) circle (0.4cm);
	\node [blue,inner sep = 0] at (6,1) {FPZ};
	\draw [fill] (6,-0.5) circle (1pt);
	\draw [fill] (6,-0.25) circle (1pt);
	\draw [fill] (6,-0.75) circle (1pt);
	\draw [orange, line width = 1pt] (6,-2) circle (0.4cm);
	\node [orange,inner sep = 0] at (6,-2) {CB2};
\draw[thick] (-0.0,2.65) -- (-0.0,-2.65);      
\draw[thick] (+0.3,2.65) -- (0,2.65);          
\draw[thick] (+0.3,-2.65) -- (0,-2.65);        

\draw[thick] (5.0,2.65) -- (5.0,-2.65);        
\draw[thick] (4.7,2.65) -- (5,2.65);           
\draw[thick] (4.7,-2.65) -- (5,-2.65);         

	\begin{scope}[yshift=-0.25cm]
		\draw [fill = red] (0.25,-4) rectangle (0.75,-3.2);
		\node [] at (0.5,-4.5) {\footnotesize Delta};
		\draw [fill = yellow] (1.25,-4) rectangle (1.75,-3.47);
		\node at (1.5,-4.5) {\footnotesize Theta};
		\draw [fill = cyan] (2.25,-4) rectangle (2.75,-3.47);
		\node at (2.5,-4.5) {\footnotesize Alpha};
		\draw [fill = ForestGreen] (3.25,-4) rectangle (3.75,-3.47);
		\node at (3.5,-4.5) {\footnotesize Beta};
		\draw [fill = Purple] (4.25,-4) rectangle (4.75,-3);
		\node at (4.5,-4.5) {\footnotesize Gamma};
		\draw[,thick, rounded corners=2mm] (0,-5)rectangle(5,-2.75);
	
		\draw[orange,dashed ,thick, rounded corners=2mm] (0.15,-2.05)rectangle(4.8,-1.5);
	
		\draw[blue,dashed ,thick, rounded corners=2mm] (0.15,+1.0)rectangle(4.8,1.5);
	
		\draw[Brown,dashed ,thick, rounded corners=2mm] (0.15,+2.0)rectangle(4.8,2.5);

	\end{scope}
	
	\draw [<-] (0.5,-3) -- (0.5,-2.5);
	\draw [<-] (1.5,-3) -- (1.5,-2.5);
	\draw [<-] (2.5,-3) -- (2.5,-2.5);
	\draw [<-] (3.5,-3) -- (3.5,-2.5);
	\draw [<-] (4.5,-3) -- (4.5,-2.5);

	\draw [->] (5,2) -- (5.5,2);
	\draw [->] (5,1) -- (5.5,1);
	\draw [->] (5,-2) -- (5.5,-2);
	
	\draw[,thick, rounded corners=2mm] (6.5,-2.75)rectangle(5.5,2.75);
	\draw[Purple, dashed, thick, rounded corners=2mm]  (4.1,-2.5)rectangle(4.9,2.5);
	 
	\draw[ForestGreen, dashed, thick, rounded corners=2mm]  (3.1,-2.5)rectangle(3.9,2.5);

	\draw[cyan, dashed, thick, rounded corners=2mm]  (2.1,-2.5)rectangle(2.9,2.5);
	
	\draw[yellow, dashed, thick, rounded corners=2mm]  (1.1,-2.5)rectangle(1.9,2.5);
	\draw[red, dashed, thick, rounded corners=2mm]  (0.1,-2.5)rectangle(0.9,2.5);
	
\end{tikzpicture}
	\caption{Visualization of EEG feature importance with respect to frequency bands and channels.}
	\label{fig:theta-matrix}
\end{figure}
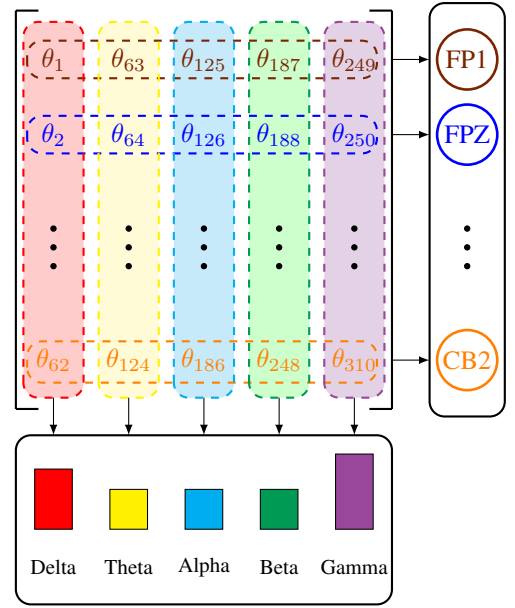

Based on the established correspondence between EEG feature dimensions and frequency bands/channels, the overall importance of the $j$-th frequency band is computed as:
\begin{equation}
	\omega^{\text{band}}_{j} = \sum_{i=1}^{q} \boldsymbol{\theta}_{(j-1)\times q+i}, \quad j = 1, 2, \ldots, p,
	\label{eq:band_importance}
\end{equation}
where $p=5$ denotes the number of frequency bands and $q=62$ represents the number of EEG channels. Similarly, the importance of the $k$-th EEG channel is measured by:
\begin{equation}
	\omega^{\text{channel}}_{k} = \sum_{i=1}^{p} \boldsymbol{\theta}_{(i-1)\times q+k}, \quad k = 1, 2, \ldots, q.
	\label{eq:channel_importance}
\end{equation}
Since the feature representation is constructed by concatenating $62$ EEG channels across $5$ frequency bands, the importance vector $\boldsymbol{\theta} \in \mathbb{R}^{310}$ can be reshaped into a matrix $\boldsymbol{\Theta} \in \mathbb{R}^{62 \times 5}$, where each column corresponds to a frequency band and each row corresponds to an EEG channel (Fig.~\ref{fig:theta-matrix}). The importance of each frequency band and \gls{eeg} channel is obtained by summing the
elements along the corresponding columns and rows, respectively.

The results shown in Fig.~\ref{fig:impBand} illustrate the average importance of each frequency band across three cross-session tasks. Among all frequency bands, the Gamma band exhibits the highest contribution to emotion recognition, followed by the Alpha band.

\begin{figure}
	\centering
	\setlength\axheight{0.5\linewidth}%
	\setlength\axwidth{0.8\linewidth}%
	\setlength\myfontsize{8pt}%
%
%
\definecolor{mycolor1}{rgb}{0.12941,0.12941,0.12941}%
\begin{tikzpicture}[font=\fontsize{\myfontsize}{1.2\myfontsize}\selectfont]
\begin{axis}[%
width=\axwidth,
height=\axheight,
at={(0in,0in)},
scale only axis,
bar shift auto,
xmin=-0.2,
xmax=11.2,
xtick={1,2,3,4,5,6,7,8,9,10},
xticklabels={{CPZ},{CZ},{FP1},{CP2},{FP2},{C2},{T8},{FPZ},{TP8},{F7}},
xlabel style={font=\fontsize{\myfontsize}{1.2\myfontsize}\selectfont, color=white!15!black},
ymin=0,
ymax=0.025,
ylabel style={font=\fontsize{\myfontsize}{1.2\myfontsize}\selectfont, color=white!15!black},
xtick pos=bottom,
scaled y ticks=false,
yticklabel style={
	/pgf/number format/.cd,
	fixed,
	precision=4
},
ytick={0,0.005,0.01,0.015,0.02,0.025},
xtick pos=bottom,
ytick pos=left,
]
\addplot[ybar, bar width=0.8, fill=black!50!blue, draw=mycolor1, area legend] table[row sep=crcr] {%
1	0.0246170344712168\\
2	0.0234497011410576\\
3	0.0226488047604331\\
4	0.0226288098852592\\
5	0.0224543279531514\\
6	0.0199830253943321\\
7	0.0197586526459224\\
8	0.0195592048656521\\
9	0.0192373798829774\\
10	0.018916046760556\\
};
\addplot[forget plot, color=mycolor1] table[row sep=crcr] {%
-0.2	0\\
11.2	0\\
};
\end{axis}
\end{tikzpicture}%
	\caption{Average EEG channel importance obtained by \gls{egda} for cross-session emotion recognition.
	 }
	\label{fig:impChannel}
\end{figure}
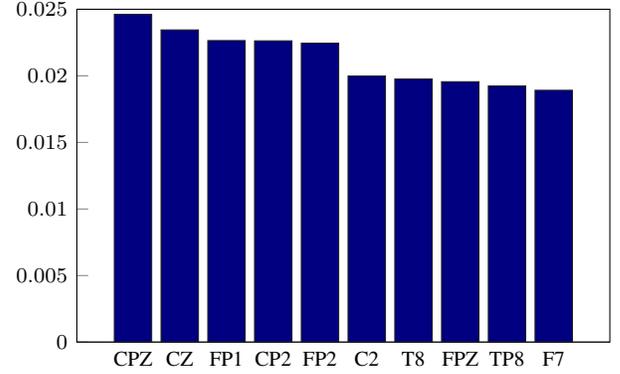

\begin{figure}
	\centering
	\setlength\axheight{0.5\linewidth}%
	\setlength\axwidth{0.5\linewidth}%
	\setlength\myfontsize{8pt}%
	\pgfplotsset{
	colormap={turbo}{
		rgb255(0cm)=(48,18,59)
		rgb255(0.13cm)=(63,69,174)
		rgb255(0.25cm)=(34,148,222)
		rgb255(0.38cm)=(49,209,162)
		rgb255(0.50cm)=(183,223,72)
		rgb255(0.63cm)=(255,200,33)
		rgb255(0.75cm)=(255,123,0)
		rgb255(0.88cm)=(200,35,1)
		rgb255(1cm)=(128,0,38)
	}
}
\begin{tikzpicture}[font=\fontsize{\myfontsize}{1.2\myfontsize}\selectfont]
	\begin{axis}[
		width=\axwidth,
		height=\axheight,
		at={(0in,0in)},
		scale only axis,
		unbounded coords=jump,
		xmin=-10,
		xmax=10,
		ymin=-10,
		ymax=10,
		axis background/.style={fill=white},
		enlargelimits=false,
		point meta min= 0.0138  ,
		point meta max= 0.0271,
		axis lines=none,
		ticks=none,
		xlabel={},
		ylabel={},
		colorbar,
	colorbar style={
		width=0.25cm,
		xshift=0cm,
		ytick={0.0138,0.0182,0.0227,0.0271},
		scaled y ticks=false,
		yticklabel style={
			/pgf/number format/.cd,
			precision=4,
		},
	}
		]
		\addplot [forget plot] graphics [xmin=-10, xmax=10, ymin=-10, ymax=10] {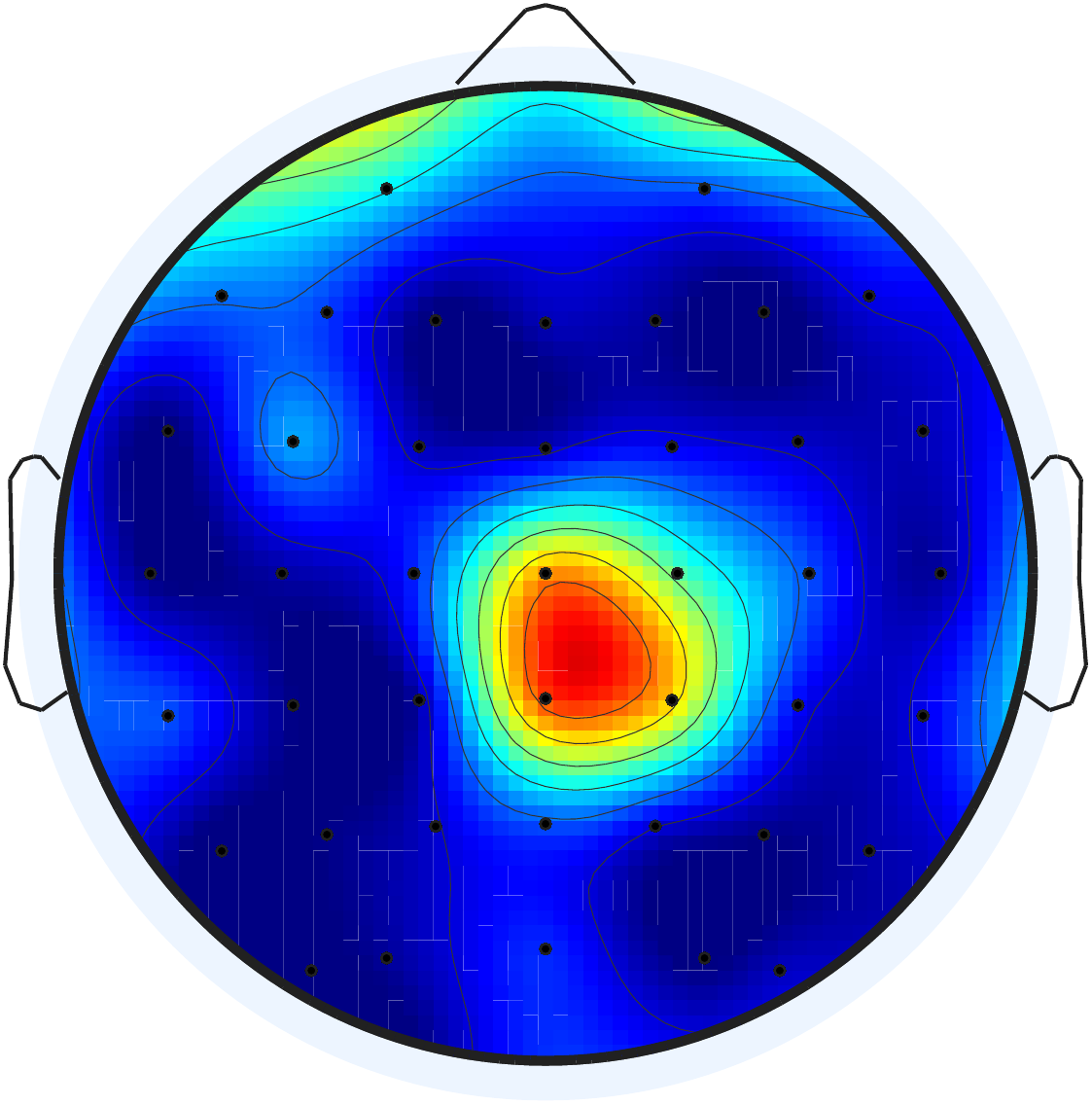};
	\end{axis}
\end{tikzpicture}
	\caption{Topographic map of the average channel importance obtained by \gls{egda} for
		cross-session EEG emotion recognition.}
	\label{fig:topo}
\end{figure}

\begin{figure}
	\centering
	\setlength\axheight{0.6\linewidth}%
	\setlength\axwidth{0.6\linewidth}%
	\setlength\myfontsize{8pt}%
	\begin{tikzpicture}[font=\fontsize{\myfontsize}{1.2\myfontsize}\selectfont]
	\begin{axis}[ 
		width=\axwidth, 
		height=\axheight, 
		at={(0in,0in)}, 
		scale only axis, 
		unbounded coords=jump, 
		xmin=-10, xmax=10, ymin=-10, ymax=10, 
		axis background/.style={fill=white}, 
		enlargelimits=false,
		point meta min=0.0138, 
		point meta max=0.0271, 
		axis lines=none, 
		ticks=none, 
		xlabel={}, ylabel={},
		legend style={at={(0.5,1.05)}, anchor=south, legend columns=1} 
		]
		
		\addplot [forget plot] graphics [xmin=-10, xmax=10, ymin=-10, ymax=10] {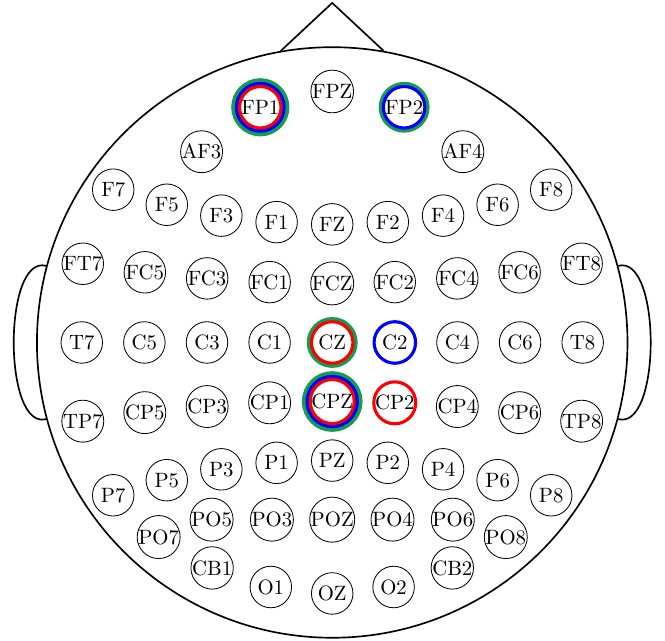}; 
		
		\addplot+[only marks, mark=o, thick, green] coordinates {(-999,-999)};
		\addlegendentry{Session 1 $\rightarrow$ Session 2}
		
		\addplot+[only marks, mark=o, thick, blue] coordinates {(-999,-999)};
		\addlegendentry{Session 1 $\rightarrow$ Session 3}
		
		\addplot+[only marks, mark=o, thick, red] coordinates {(-999,-999)};
		\addlegendentry{Session 2 $\rightarrow$ Session 3}

	\end{axis}
\end{tikzpicture}
	\caption{The four most important EEG channels selected for each cross-session task.}
		\label{fig:channel_selection}
\end{figure}
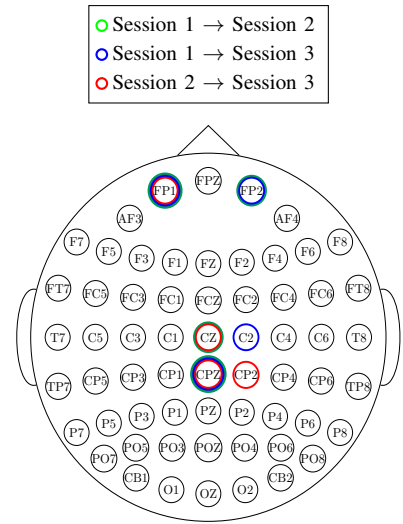

At the channel level, Fig.~\ref{fig:impChannel} presents the ten most important EEG channels averaged across the three cross-session tasks. These channels are primarily distributed over the prefrontal, central-parietal, and bilateral temporal regions and exhibit the highest discriminative contributions, as illustrated in Fig.~\ref{fig:topo}.
 Furthermore, Fig.~\ref{fig:channel_selection} highlights four EEG channels that are consistently identified as the most discriminative across all trials. These findings are consistent with previous studies, confirming the critical role of these brain regions in emotional processing \cite{zhang2020selection}.

\begin{figure}
	\centering
	\setlength\axheight{0.5\linewidth}%
	\setlength\axwidth{0.8\linewidth}%
	\setlength\myfontsize{8pt}%
%
%
\definecolor{mycolor1}{rgb}{0.12941,0.12941,0.12941}%
\begin{tikzpicture}[font=\fontsize{\myfontsize}{1.2\myfontsize}\selectfont]
	\begin{axis}[%
		width=\axwidth,
		height=\axheight,
		at={(0in,0in)},
scale only axis,
bar shift auto,
xmin=-0.2,
xmax=6.2,
xtick={1,2,3,4,5},
xticklabels={{Delta},{Theta},{Alpha},{Beta},{Gamma}},
xlabel style={font=\fontsize{\myfontsize}{1.2\myfontsize}\selectfont, color=white!15!black},
xlabel={},
ymin=0,
ymax=0.282484762179957,
ylabel style={font=\fontsize{\myfontsize}{1.2\myfontsize}\selectfont, color=white!15!black},
ylabel={},
axis background/.style={fill=white},
xtick pos=bottom,
scaled y ticks=false,
yticklabel style={
	/pgf/number format/.cd,
	fixed,
	precision=4
},
xtick pos=bottom,
scaled y ticks=false,
yticklabel style={
	/pgf/number format/.cd,
	fixed,
	precision=4
},
ytick={0,0.05,0.1,0.15,0.2,0.25,0.3,0.35},
xtick pos=bottom,
ytick pos=left,
]
\addplot[ybar, bar width=0.8, fill=black!50!blue, draw=mycolor1, area legend] table[row sep=crcr] {%
1	0.213408952272555\\
2	0.18534463872656\\
3	0.185673344035837\\
4	0.18308830278509\\
5	0.232484762179957\\
};
\addplot[forget plot, color=mycolor1] table[row sep=crcr] {%
-0.2	0\\
6.2	0\\
};
\node[align=center, inner sep=0, font=\fontsize{1.1\myfontsize}{1.32\myfontsize}\selectfont]
at (axis cs:1,0.233) {0.2134};
\node[align=center, inner sep=0, font=\fontsize{1.1\myfontsize}{1.32\myfontsize}\selectfont]
at (axis cs:2,0.205) {0.1853};
\node[align=center, inner sep=0, font=\fontsize{1.1\myfontsize}{1.32\myfontsize}\selectfont]
at (axis cs:3,0.206) {0.1857};
\node[align=center, inner sep=0, font=\fontsize{1.1\myfontsize}{1.32\myfontsize}\selectfont]
at (axis cs:4,0.203) {0.1831};
\node[align=center, inner sep=0, font=\fontsize{1.1\myfontsize}{1.32\myfontsize}\selectfont]
at (axis cs:5,0.252) {0.2325};
\end{axis}
\end{tikzpicture}%
	\caption{Average importance of EEG frequency bands obtained by \gls{egda} for cross-session emotion recognition
	 }
	\label{fig:impBand}
\end{figure}
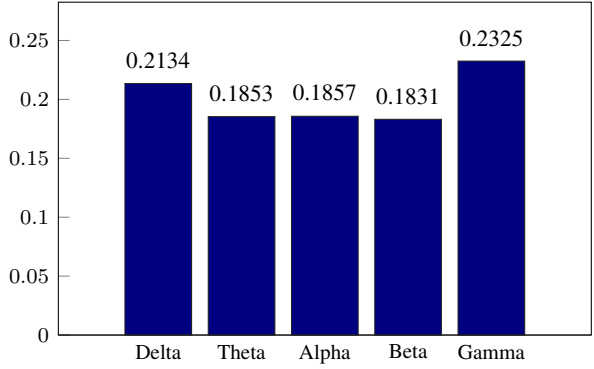

%
%
%
%
%

\section{Conclusion}

In this work, we proposed the \gls{egda} framework for robust cross-session emotion recognition. The proposed method integrates distribution alignment with graph regularization to simultaneously reduce marginal and conditional discrepancies across sessions. By minimizing within-class scatter matrices in the source domain, \gls{egda} encourages compact and discriminative feature representations, thereby facilitating more effective domain adaptation. Experimental results on the SEED-IV dataset demonstrate that \gls{egda} achieves superior recognition accuracy compared with several state-of-the-art baseline methods. Furthermore, our analysis highlights the Gamma frequency band as the most discriminative and identifies the prefrontal, central-parietal, and temporal brain regions as critical for emotional processing.

\bibliographystyle{ieeetr}

\end{document}